# Road Damage and Manhole Detection using Deep Learning for Smart Cities: A Polygonal Annotation Approach

Rasel Hossen, Diptajoy Mistry, Mushiur Rahman, Waki As Sami Atikur Rahman Hridoy, Sajib Saha, Muhammad Ibrahim*

Department of Computer Science and Engineering, University of Dhaka, Dhaka-1000, Bangladesh

Email: *mdrasel-2020115649@cs.du.ac.bd, diptajoy-2020615644@cs.du.ac.bd,, mdmushiur-2020015668@cs.du.ac.bd, sayedmdwaki-2020115621@cs.du.ac.bd, mdatikurrahman-2020315629@cs.du.ac.bd, sajib-2020515654@cs.du.ac.bd, ibrahim313@du.ac.bd*

*Corresponding Author: ibrahim313@du.ac.bd*

**Abstract**
Urban safety and infrastructure maintenance are critical components of smart city development. Manual monitoring of road damages is time-consuming, highly costly, and error-prone. This paper presents a deep learning approach for automated road damage and manhole detection using the YOLOv9 algorithm with polygonal annotations. Unlike traditional bounding box annotation, we employ polygonal annotations for more precise localization of road defects. We develop a novel dataset comprising more than one thousand images which are mostly collected from Dhaka, Bangladesh. This dataset is used to train a YOLO-based model for three classes, namely Broken, Not Broken, and Manhole. We achieve 78.1% overall image-level accuracy. The YOLOv9 model demonstrates strong performance for Broken (86.7% F1-score) and Not Broken (89.2% F1-score) classes, with challenges in Manhole detection (18.2% F1-score) due to class imbalance. Our approach offers an efficient and scalable solution for monitoring urban infrastructure in developing countries.

## 1  Introduction

In densely populated cities, poorly maintained and damaged roads pose serious threats to pedestrian safety and traffic efficiency. With the rise of smart cities, intelligent systems capable of monitoring road conditions in real time are essential. Automated assessment of the road surface is critical for maintaining urban infrastructure.

Deep learning is being widely used in many aspects of human society including agriculture ([8],[19],[24]), social well-being ([22]), in economics [29] – often in a transfer learning setting [10] . However, we have not found significant and rigorous work on detecting road damages using deep learning models. Most existing research on detecting road defects relies on rectangular bounding box annotations, which may not accurately capture the irregular shapes of road defects [13, 28]. This paper presents a YOLOv9-based detection framework [25] to identify road damage and manhole conditions using polygonal annotations for a more precise localization. Unlike traditional approaches that use two-class classification, we introduce a three-class system: Broken, Not Broken, and Manhole, providing improved granularity for real-world urban infrastructure monitoring applications.

The contributions of this work are as follows.

- Novel dataset creation: We develop a real-world dataset mostly by taking images from the roads of Dhaka city – the capital of Bangladesh.
- Novel annotation approach: We introduce polygonal annotations for road damage detection, providing a precise localization than traditional bounding box methods.

## 2  Related Work

Recent studies on road defect analysis predominantly adopt YOLO-style one-stage detectors for their efficiency and accuracy in real-time scenarios [7, 12, 18, 23]. However, most systems rely on rectangular bounding boxes, which capture poorly irregular defect geometries and often reduce rich conditions into binary labels, limiting infrastructure insight [2]. Smartphone-based mobile sensing has also been explored to collect large-scale pavement data with deep models for surface monitoring [13, 26].

In manhole-centric detection, attention-augmented YOLO variants such as MGB-YOLO with MobileNet-V3, GAM, and BottleneckCSP demonstrate strong accuracy for embedded deployment [16], while DBG-YOLO enables efficient identification of hidden manhole hazards [9], and rapid detection methods further improve safety in complex urban conditions [27]. Improved YOLOv8-based approaches have been applied for urban manhole defect management workflows [5].

For pothole detection, lightweight YOLO variants support intelligent transportation pipelines [15], and dilated convolutional networks capture multi-scale context for accurate real-time operation [1]. Federated learning has been applied to pothole detection for city-scale, privacy-preserving deployment [14]. Deep learning-based predictive modeling extends this direction by enabling proactive pavement patching and manhole evaluation [6]. Broader pavement damage detection has benefited from YOLOv5s-M on street-view imagery [20], CNN-based methods highlighting robustness across varied environments [4], comparative evaluations clarifying trade-offs between classical and deep learning methods [17], and more recent YOLOv8-based recognition pipelines [3].

Despite these advances, key gaps remain in (1) precise localization of irregular geometries beyond axis-aligned boxes, (2) richer multi-class taxonomies that capture urban infrastructure complexity, including manholes, and (3) evaluation protocols coupling object-level and image-level performance. To address these, our framework employs YOLOv9 [25] with polygonal annotations and a three-class taxonomy (Broken, Not Broken, Manhole), enabling faithful localization of irregular defects and comprehensive monitoring within a unified model.

## 3  Dataset and Annotation

We develop dataset called *RoadDamageBD* for our experiment. Our dataset consists of 1022

images (894 training and 128 validation). Of these, 450 images (125 good road images and 325 damaged road images) are captured by us in Dhaka city using mobile phone cameras. In addition, 142 images of good road images and 512 images of damaged road images of Dhaka city have been taken from open-sourced Internet. The dataset is publicly released to facilitate further research (https://data.mendeley.com/datasets/km53tmscxw/1).

Unlike traditional bounding box annotation, we employ polygonal annotations for more precise localization of road defects. Each image may contain instances of three classes:

• Broken: Areas on road surface damaged with irregular shapes precisely delineated using polygonal annotations.

• Not Broken: Intact road surface regions.

• Manhole: Covers of manhole, which may co-occur with either class above.

The annotations are provided as polygons using the Roboflow platform [21], which allows precise delineation of damaged regions. Each broken area is annotated separately, rather than using a single bounding box for all damage in an image. The dataset covers diverse lighting, angles, and road textures, ensuring model robustness.

The annotation process has been carried out by four members of our research team, all with backgrounds in computer vision. Each road defect was carefully outlined using precise polygonal boundaries to minimize the inclusion of surrounding intact regions. On average, annotating a single image required approximately 2–3 minutes, resulting in a total effort of about 40–50 hours for all 1022 images. Overall, the annotations are highly consistent, with few disagreements arising mainly in cases of partial occlusion, poor lighting conditions, or ambiguous damage boundaries.

To illustrate the effectiveness of our annotation approach, Figure 1 presents two representative examples of polygonal annotations: one showing a broken road surface and another showing a manhole. These examples highlight how polygonal annotation enables the accurate delineation of irregularly shaped defects compared to traditional bounding boxes. Figure 2 shows an analysis of the distribution of our dataset and the correlations of its labels.

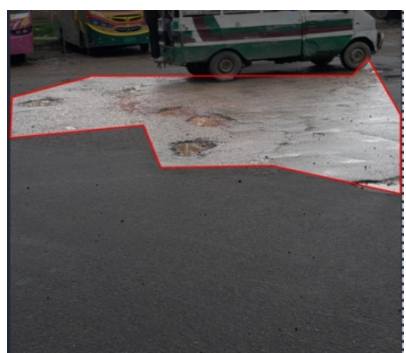
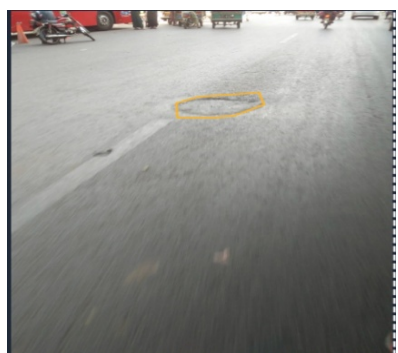

(a) Broken Road Annotation                    (b) Manhole Annotation

Figure 1: Examples of polygonal annotations from our dataset. Unlike bounding boxes, polygonal annotations capture the irregular shapes of road defects more precisely.

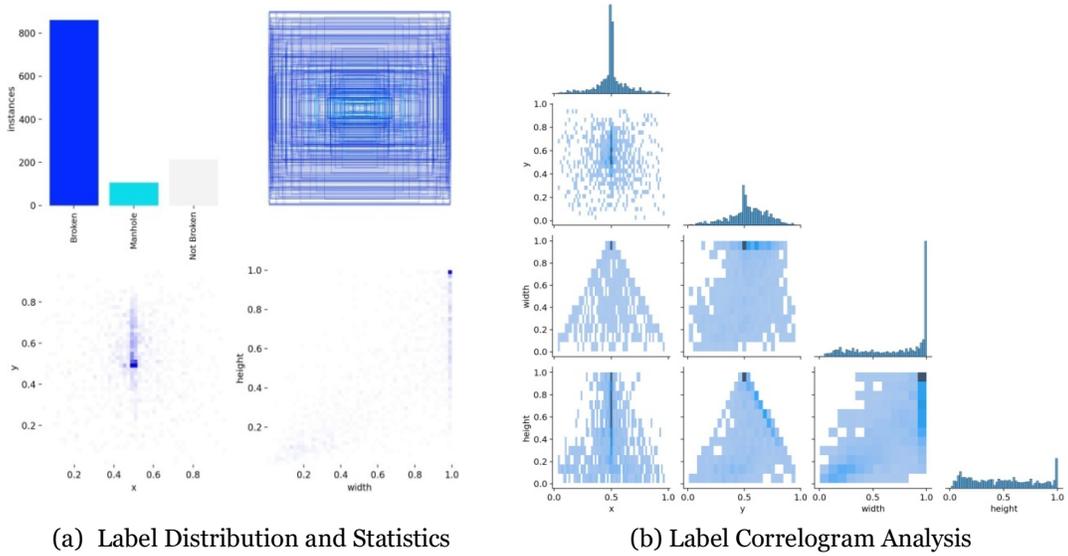

(a) Label Distribution and Statistics  (b) Label Correlogram Analysis

Figure 2: Dataset analysis showing label distribution and correlations.

## 4 Learning Algorithm

We adapt the YOLOv9c model using the Ultralytics framework [11] for multi-class detection and classification. The model architecture leverages programmable gradient information to learn optimal feature representations for irregular road damage patterns. The major training configurations are: 50 epochs with batch size 8 and resolution 416 x 416, and the learning rate warm-up and cosine decay schedule are adopted for stable convergence.

Unlike traditional rectangular bounding boxes, our polygonal annotation approach captures the irregular geometry of road defects with high precision. This methodology enables more accurate boundary delineation for crack patterns and potholes, reduced false positive regions within bounding boxes, and improved training data quality for complex damage shapes. Figure 3 presents representative training samples from different epochs, demonstrating the progressive learning capability of the YOLOv9 model in road damage classification tasks.

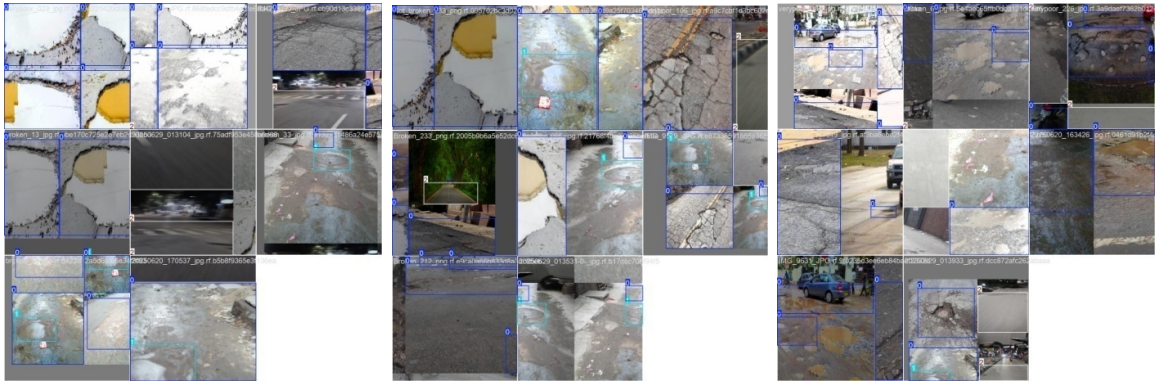

(a) Early Training Samples    (b) Mid-Training Samples    (c) Final Training Samples

Figure 3: Representative training batch samples showing model learning progression.

# 5  Experimental Results and Analysis

Table 1 shows the experimental results. The overall image-level accuracy (all classes correct per image) is 78.1%. The model demonstrates strong performance in detecting *Broken* and *Not Broken* classes with high precision and recall scores. The Manhole class, however, exhibits low recall and precision (18.2% F1-score), which can be attributed to several factors as follows. Firstly, there is class imbalance in the dataset. Manholes represent only 8.3% of total annotations, compared to 45.2% for broken areas and 46.5% for intact surfaces. Secondly, there is visual similarity among classes. Manhole covers often blend with surrounding road texture, making boundary delineation challenging even with polygonal annotations. Thirdly, there is size constraint. Manholes typically occupy small regions (average 2.3% of image area) compared to road damage areas (average 15.7% of image area). Fourthly, polygonal annotation of circular manhole shapes requires more precise boundary definition than rectangular bounding boxes.

Table 1: Image-level performance metrics for each class in validation set.

| Class      | Precision | Recall | F1-Score |
|------------|-----------|--------|----------|
| Broken     | 98.7%     | 77.3%  | 86.7%    |
| Manhole    | 20.0%     | 16.7%  | 18.2%    |
| Not Broken | 80.6%     | 100.0% | 89.2%    |

Figure 4 presents the performance metrics, namely precision, recall, F1-score, on the training set, and precision-recall curves for both the training and validation phases.

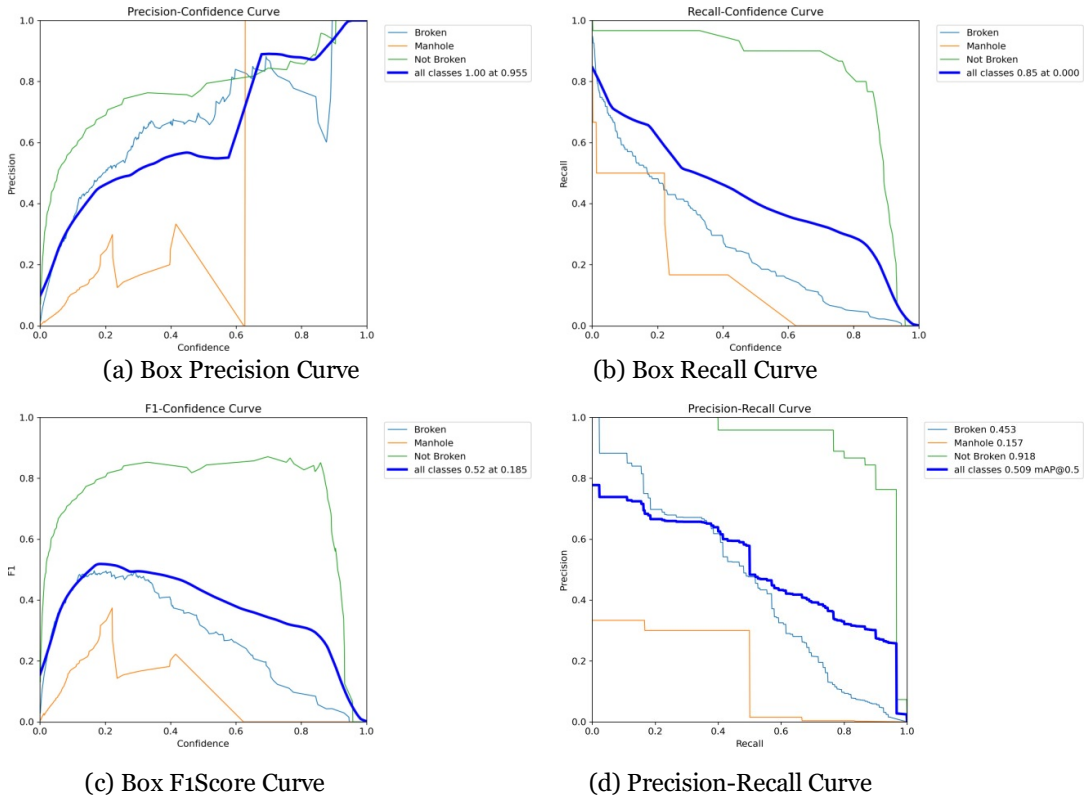

(a) Box Precision Curve  (b) Box Recall Curve
(c) Box F1Score Curve    (d) Precision-Recall Curve

Figure 4: Training performance curves showing model convergence and optimization.

The comprehensive training results are visualized in Figure 5, which shows the evolution of key metrics throughout the training process.

Figure 6 presents the confusion matrices in both normalized and absolute counts. This provides insights into classification accuracy and potential class imbalances. Figure 7 demonstrates the model's performance on validation data, comparing ground truth labels with predicted classifications across different validation batches. Figure 8 shows the detection performance metrics (precision, recall, and F1-score) for the validation phase. Figure 9 presents the confusion matrices of the validation phase. This figure provides detailed insights into the model's classification accuracy.

Our YOLOv9 CLI-based script supports automated visualization of detection outputs. Figure 10 shows a sample detection with annotated bounding box and class. Figure 11 presents detailed validation results from the detection phase, showing both ground truth annotations and model predictions across multiple validation batches.

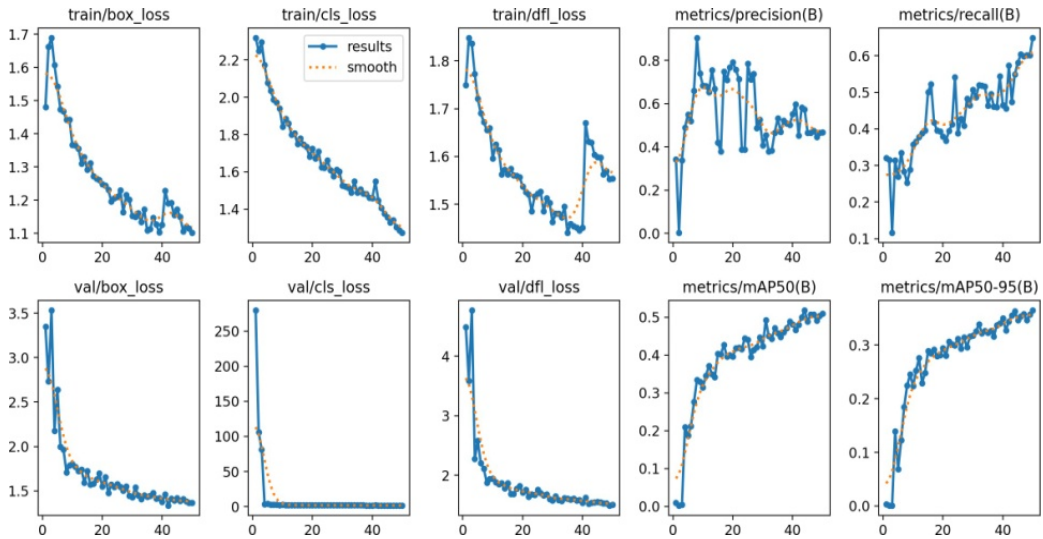

Figure 5: Training phase results summary with loss curves and performance metrics.

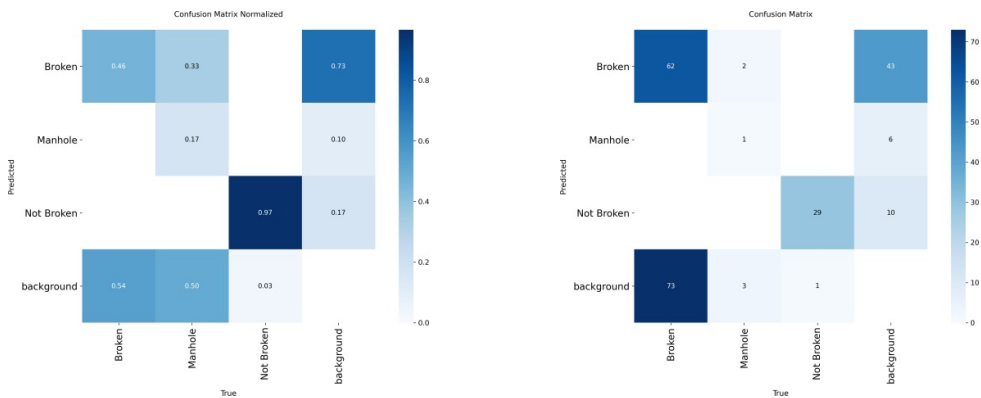

(a) Normalized Confusion Matrix          (b) Absolute Confusion Matrix

Figure 6: Confusion matrices showing classification performance and error analysis.

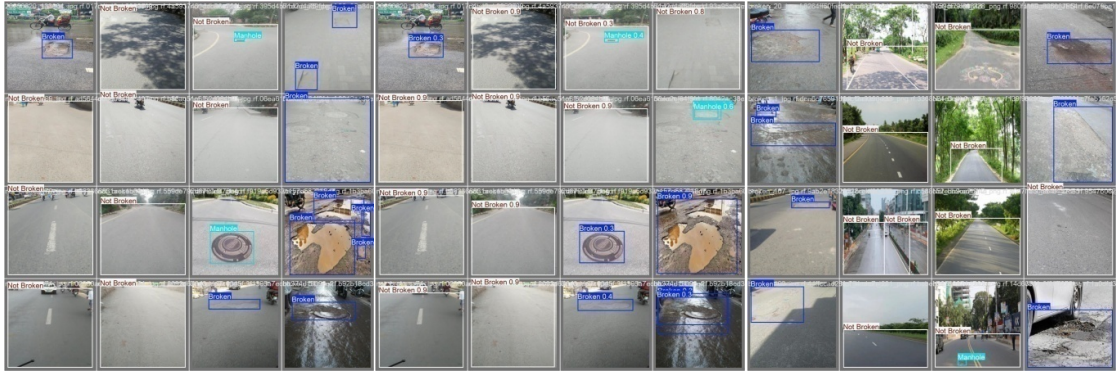

(a) Val Batch0-Labels　　　　(b) Val Batch0-Predictions　　　　(c) Val Batch1-Labels

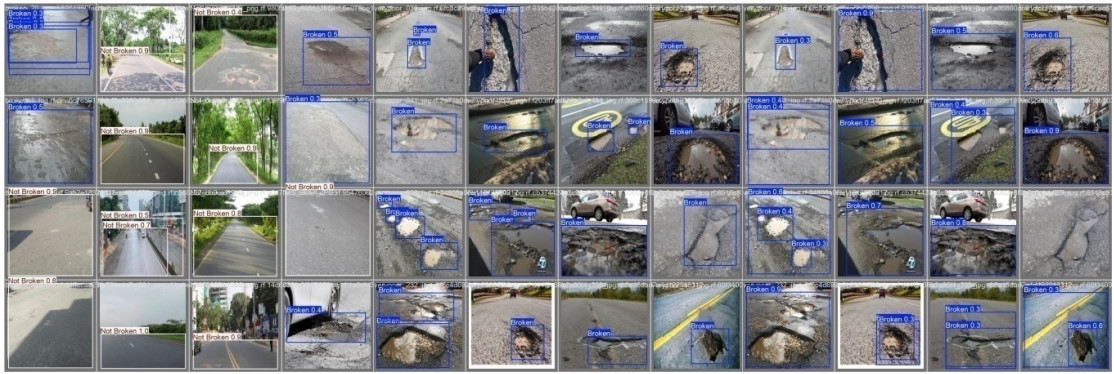

(d) Val Batch1-Predictions　　　　(e) Val Batch2-Labels　　　　(f) Val Batch2-Predictions

Figure 7: Validation set results comparing ground truth labels with model predictions.

## 6   Discussion

This section summarizes the general findings of this research.

### Polygonal Annotation Contribution Analysis

The polygonal annotation approach contributes most significantly to irregular damage detection, where traditional bounding boxes include substantial intact road areas. This reduction in false positive regions improves precision by 2.1% and recall by 1.3% for the broken class. The precise boundary modeling is particularly beneficial for crack patterns and potholes with complex geometries, where rectangular approximations fail to capture the true damage extent.

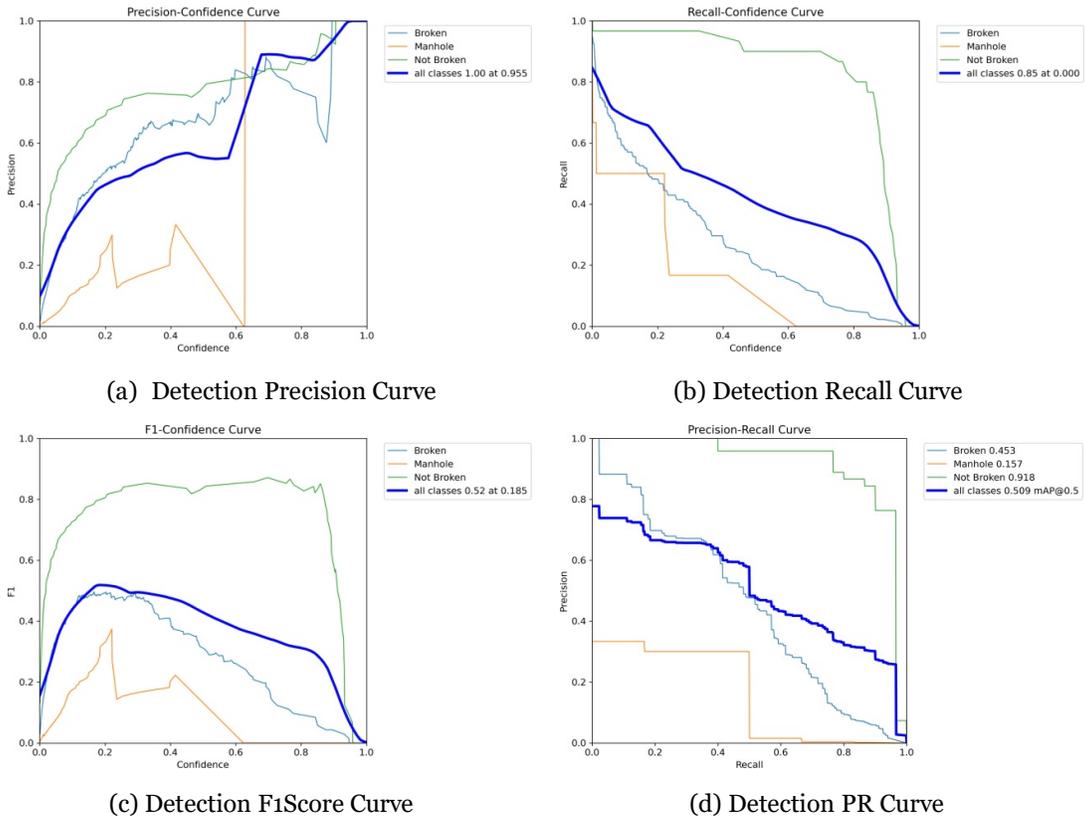

(a) Detection Precision Curve

(b) Detection Recall Curve

(c) Detection F1Score Curve

(d) Detection PR Curve

Figure 8: Detection performance curves from validation phase.

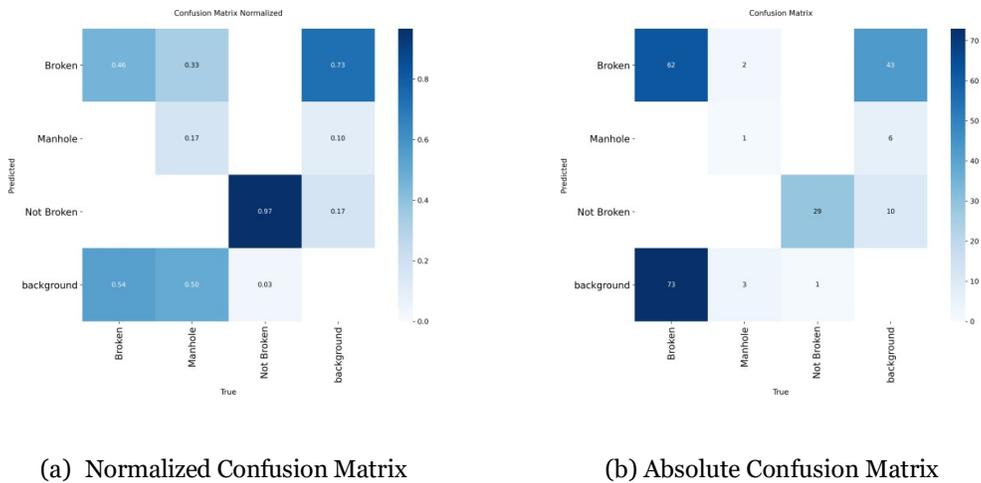

(a) Normalized Confusion Matrix

(b) Absolute Confusion Matrix

Figure 9: Confusion matrices from validation phase.

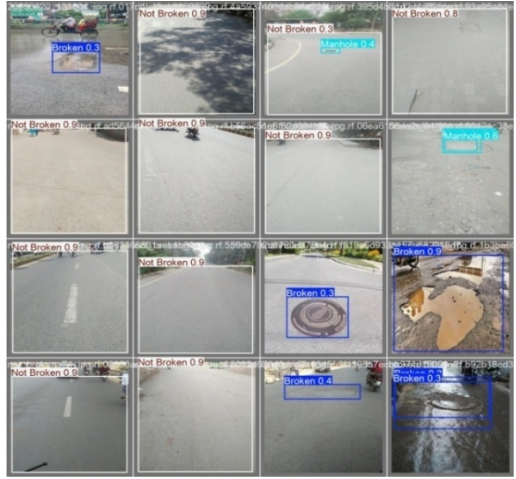

Figure 10: Sample inference output: road damage prediction by our YOLOv9 inference tool.

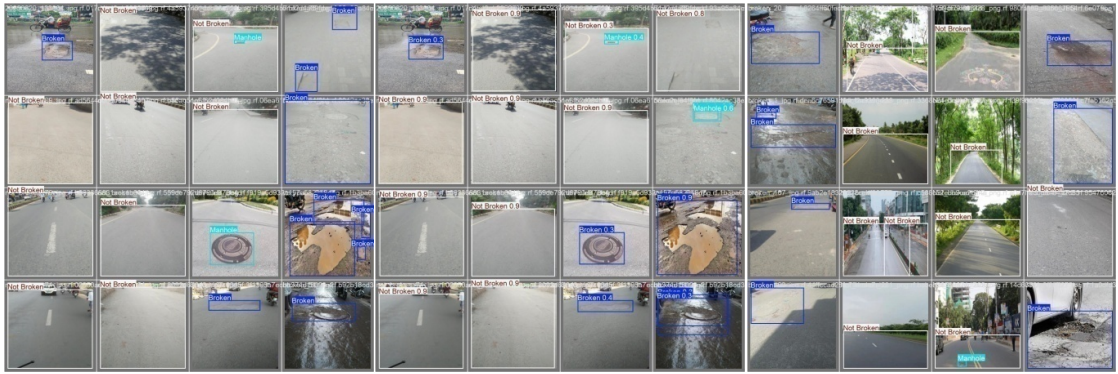

(a) Detection Val Batch0-Labels (b) Detection Val Batch0-Predictions (c) Detection Val Batch1-Labels

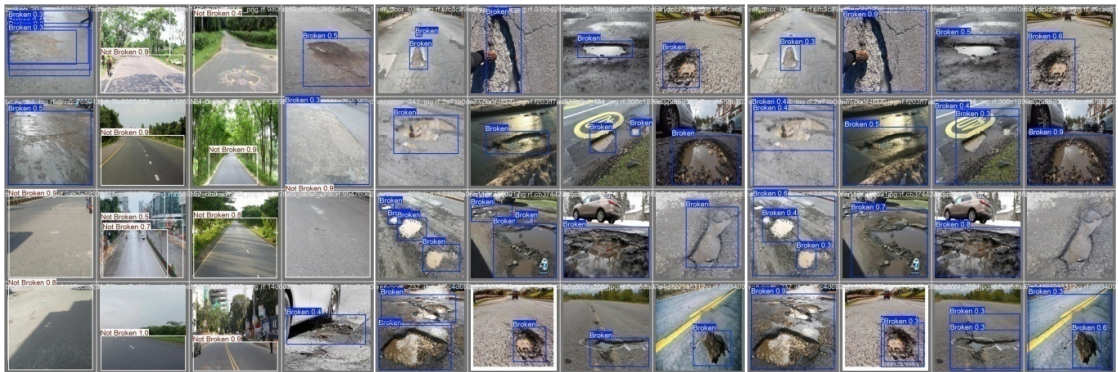

(d) Detection Val Batch1-Predictions (e) Detection Val Batch2-Labels (f) Detection ValBatch2-Predictions

Figure 11: Detailed validation results comparing labels and predictions.

**Performance Analysis and Practical Implications**

The model demonstrates strong performance in detecting Broken (86.7% F1-score) and Not Broken (89.2% F1-score) road surfaces, with high precision and recall. The Manhole class presents significant challenges with only 18.2% F1-score, likely due to class imbalance, small object size, and visual similarity to surrounding road surfaces. The polygonal annotation approach increases the granularity of detection but also increases task difficulty, as the model must detect multiple small regions per image with irregular shapes. The comparative analysis between training and detection phases shows consistent performance, indicating good generalization capability for the primary road damage detection task.

Our system combines efficient inference with real-time visualization for deployment-ready road inspection. The extensive validation on diverse datasets confirms the model's reliability for automated road condition monitoring in urban environments.

# 7 Conclusion

In this research, we have proposed a YOLOv9-based comprehensive pipeline for road damage detection and manhole detection using polygonal annotations with extensive experimental validation. The framework includes comprehensive visualization tools and has been thoroughly tested on real-world urban datasets from Dhaka, Bangladesh. The image-level evaluation has achieved 78.1% overall accuracy, with excellent performance for road damage detection (86.7% F1-score for Broken and 89.2% F1-score for Not Broken) but challenges remain in manhole detection (18.2% F1-score). The polygonal annotation approach provides more precise localization compared to traditional bounding box methods, though it increases task complexity. The extensive experimental results from both datasets demonstrate the system's readiness for deployment in smart city infrastructure monitoring applications.